\documentclass[conference]{IEEEtran}

\usepackage{graphicx}
\usepackage{multirow}

\usepackage{booktabs}
\usepackage{csquotes}
\usepackage{tabularx}
\usepackage{amssymb}
\usepackage{lipsum}
\usepackage{multirow}
\usepackage{graphicx}
\usepackage{wrapfig}
\usepackage{amsmath,amsfonts,bm}
\usepackage{amsthm}
\usepackage{bm}
\usepackage{dsfont}
\usepackage{enumitem}

\usepackage{url}
\usepackage{graphicx}
\def\eqref#1{equation~\ref{#1}}
\def\1{\bm{1}}

\DeclareMathAlphabet{\mathsfit}{\encodingdefault}{\sfdefault}{m}{sl}
\SetMathAlphabet{\mathsfit}{bold}{\encodingdefault}{\sfdefault}{bx}{n}

\newcommand{\norm}[1]{\left\lVert #1 \right\rVert}

\usepackage{cite}
\usepackage{mdwmath}
\usepackage{mdwtab}
\usepackage{eqparbox}
\usepackage[tight,footnotesize]{subfigure}
\usepackage{booktabs}
\usepackage{url}
\usepackage[colorlinks]{hyperref}

\newcommand{\header}[1]{\noindent\textbf{#1}.}

\ifCLASSINFOpdf
\else
\fi
\begin{document}
\title{Securing Biomedical Images from Unauthorized Training with Anti-Learning Perturbation}

\author{\IEEEauthorblockN{Yixin Liu\thanks{\IEEEauthorrefmark{1}The first two authors contributed equally.}\IEEEauthorrefmark{1}\textsuperscript{,\IEEEauthorrefmark{2}},
Haohui Ye\IEEEauthorrefmark{1}\textsuperscript{,\IEEEauthorrefmark{3}},
Kai Zhang\IEEEauthorrefmark{2} and
Lichao Sun\IEEEauthorrefmark{2}}
\IEEEauthorblockA{\IEEEauthorrefmark{2}
Lehigh University, Bethlehem, PA, USA}
\IEEEauthorblockA{\IEEEauthorrefmark{3}
South China University of Technology, Guangdong, China\\
\{yila22, kaz321, lis221\}@lehigh.edu, sehaohuiye@mail.scut.edu.cn}
}
\IEEEoverridecommandlockouts
\def\yixin{\textcolor{red}}

\IEEEoverridecommandlockouts

\maketitle

\begin{abstract}
The volume of open-source biomedical data has been essential to the development of various spheres of the healthcare community since more `free' data can provide individual researchers more chances to contribute. However, institutions often hesitate to share their data with the public due to the risk of data exploitation by unauthorized third parties for another commercial usage (e.g., training AI models). This phenomenon might hinder the development of the whole healthcare research community. To address this concern, we propose a novel approach termed `unlearnable biomedical image' for protecting biomedical data by injecting imperceptible but delusive noises into the data, making them unexploitable for AI models. 
We formulate the problem as a bi-level optimization and propose three kinds of anti-learning perturbation generation approaches to solve the problem. 
% The noise injected into the data is imperceptible from human visual perception, which can guarantee normal data utility for other purposes except for model traninig. 
% While this "unlearnable" technique holds promise, current methods are not effective on biomedical data due to their sparse characteristics. To overcome this challenge, we introduce the Gradient-guided Causal Masking (GGCM) module, which improves data protection effectiveness and efficiency by identifying and modifying the top-K essential pixels using gradient information. 
% We empirically verify the effectiveness of our method on multiple segmentation and classification datasets. Notably, with just a single pixel modification, we successfully decreased the accuracy of the model from the original 90\% to 10\% on the OrganSMNIST dataset. 
Our method is an important step toward encouraging more institutions to contribute their data for the long-term development of the research community.
\end{abstract}


% \fi
%%\IEEEpeerreviewmaketitle

\section{Introduction}
\label{sec: intro}
% Data important for medical community development. However, there are some  “free” explorations of personal data for unauthorized or even illegal purposes. 
The proliferation of open-source biomedical data has played a crucial role in advancing multiple aspects of the healthcare industry \cite{kostkova2016owns}. With more data available, individual researchers are provided with more opportunities to make meaningful contributions to the community. Despite this, many institutions are hesitant to share their data with the public due to concerns about unauthorized third parties using the data for commercial gains, such as training AI models \cite{carlini2021extracting, zhou2023comprehensive}. This reluctance to share data can greatly impede the progress of the entire healthcare research community. Nevertheless, from the perspective of the data owner, it is inevitable to consider the ethical implications of data sharing, and potential harm to each individual whose data is used, such as privacy violations and lack of control over data usage. Therefore, to migrate such a conflict, it is crucial to develop methods for protecting sensitive biomedical data from unauthorized AI model training while still allowing for its utility for other normal purposes, such as assistance in decision-making in diagnosis \cite{chexpert}.

\begin{figure}[thbp]
    \centering
    \includegraphics[width=\linewidth]{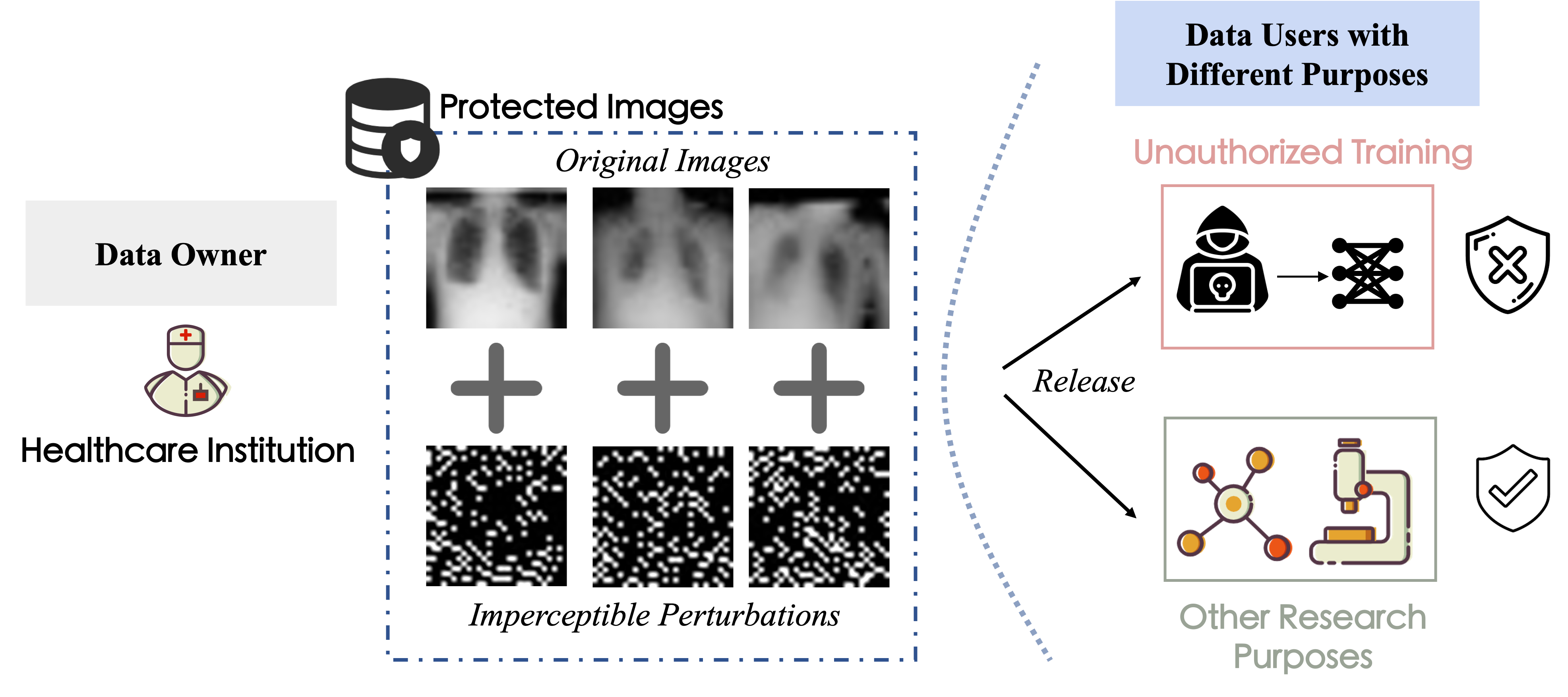}
    \caption{An illustration of motivation of protecting biomedical images from unauthorized training through the data owner's perspective. Institutions will more likely to share their data with the community if there is some guarantee that their data are free from the threat of unauthorized abuse to train models.}
    \label{fig:motivation}
\end{figure}

% To encourage.... we ask whether it is possible to prevent unauthorized training. In this paper, we address this concern by introducing unlearnable examples.

% To address this concern, which will greatly encourage more institutions to contribute their data to the community, in this paper, we propose a novel approach termed \textit{Anti-learning Perturbation} for protecting biomedical data by injecting imperceptible but delusive noises into the data, making them unexploitable and unlearnable for AI models. Specifically, we formulate the problem as a bi-level optimization and propose three kinds of anti-learning perturbation approaches to approximately solve the problem. 
In this paper, we present a new technique called `Anti-Learning Perturbation' that aims to secure biomedical data from unauthorized training by injecting imperceptible but delusive noises. This noise makes it difficult for AI models to exploit or learn from the data, thereby increasing the likelihood that institutions will be willing to share their data with the community. To accomplish this, we formulate the problem as a bi-level optimization problem and propose three distinct methods for approximately solving it. The \textit{Anti-Learning Perturbations} is designed to be imperceptible from human visual perception yet has been demonstrated to be effective in data protection, which persevers that the normal data utility for other purposes. We seek to inject misleading high-frequent signals into the training data to trick the models into relying on those brittle and inaccurate features.
% These methods are designed to be imperceptible to human observers yet have been demonstrated to be effective in data protection. 

% explain briefly about the understanding and motivation, some detail design

\section{Problem Statement}
\label{sec: ps}
% problem statement with few equations
The problem can be illustrated as a two-player game, which includes a data owner ${U}$ and an unauthorized user $\mathcal{A}$. Given a clean training dataset $\mathcal{D}^c=\{x_i,y_i\}_{i=1}^{N}$ and testing dataset $\mathcal{D}^t$, the data owner ${U}$ seeks to protect their data by adding perturbation $\mathcal{P}^{u} =\left\{ \delta _{i}^{u}|\left\| {\delta ^u}_i \right\| _p\le \epsilon _u, \text{ for all } i=1\cdots n \right\}$ to data so that the test accuracy of the trained model on $\mathcal{D}^t$ will be decreased. And we denote the derived unlearnable version of the training dataset as $\mathcal{D}^u$. We assume that the data owner ${U}$ has full access to the biomedical data, and can do any modification with the \textit{features} of data within certain kinds of budget before the data release. After publishing the data, the data owner could not interfere with the model selection and the training procedure of the unauthorized users $\mathcal{A}$. Formally, given a classifier $f$, and the cross-entropy loss $\mathcal{L}(\cdot, \cdot)$, this task can be formalized into the following bi-level optimization problem: $\max _{\norm{\delta^{u}_i}_p \le \epsilon_u} \underset{(x, y) \sim \mathcal{D}^t}{\mathbb{E}}\left[\mathcal{L}\left(f^{*}(x), y\right)\right], \text { s.t. } f^{*} \in \underset{f}{\arg \min } \sum_{\left(x_{i}, y_{i}\right) \in \mathcal{D}^c}\left[\mathcal{L}\left(f\left(x_{i}+\delta^{u}_{i}\right), y_{i}\right)\right]$.

\section{Proposed Methodology}
\label{sec: method}
% propose three methods: tap, em, shortcut
% describe respectively
% The problem in \autoref{eq:rob unl} is hard to optimize directly due to the unknown training dynamic of DNN models. 
Directly solving the bi-level optimization is intractable for  neural networks as it requires unrolling the entire training procedure found in the inner objective (solving the optimal $f^*$) and backpropagating through it to perform a single step of gradient descent on the outer objective. Thus, the data protector $U$ must approximate the bi-level objective, which should involve some sort of heuristics. We propose three kinds of perturbation strategies: \textit{Synthetic Perturbation, Adversarial Targeted (AdvT) Perturbation, Error-Minimizing (EM) Perturbation}.

\header{Synthetic Perturbation} Intuitively, one of the most naive approaches is to inject class-wise linearly separable patterns into the image, which aims to trick the model into learning a strong correlation between the noise and the labels. To be more specific, we create a specific random patch for each class, which tricks the model into relying on these brittle patterns. 
\\
\header{Adversarial Targeted Perturbation} Another intuitive solution is to leverage a clean model to serve as a target model and avert the noise generation into a more simple adversarial example problem. Furthermore, we optimize the noise generation with the \emph{class targeted} adversarial attack.
% Specifically, we optimize the following objective: 
 
%  \begin{gather}\label{eq:adv_objective}
%      \max_{\delta \in \mathcal{P}^{u} } \,\, \bigg[ \sum_{(x_i,y_i) \in \mathcal{T}} \mathcal{L} \left( F(x_i + \delta_i; f^u), y_i \right) \bigg], 
% \end{gather}
%  where $f^u$ denotes a model trained on \emph{clean} data, which is fixed during poison generation. We call this model the \emph{crafting} model. Fittingly, this method can be called \emph{adversarial poisoning}. Furthermore, we also optimize a variant of this objective which defines a \emph{class targeted} adversarial attack. Given $g$ is a permutation (with no fixed points) on the label space, this modified objective is defined by:  
%  \begin{gather}\label{eq:targeted_objective}
%      \min_{\delta \in \mathcal{P}^{u} } \,\, \bigg[ \sum_{(x_i,y_i) \in \mathcal{T}} \mathcal{L} \left( F(x_i + \delta_i; \theta^*), g(y_i) \right) \bigg],
% \end{gather}

\header{Error-Minimizing Perturbation} Following \cite{huang2021unlearnable}, we propose a novel min-min optimization to first learn a noise generator and leverage it to conduct noise generation. The min-min optimization is solved by iteratively crafting noises that can trick the models trained on the poisoned data. 
\begin{equation}
{
\underset{\theta}{\mathrm{arg}\min}\mathbb{E} _{(x,y)\sim \mathcal{D} _c}\left[ \min_{\delta} \mathcal{L} \left( f^{\prime}(x+\delta ),y \right) \right] , \mathrm{s}.\mathrm{t}. \parallel \delta \parallel _p\le \epsilon 
\label{eq:min-min}
}
\end{equation}
% where, $f'$ denotes the source model used for noise generation. Note that this is a min-min bi-level optimization problem: the inner minimization is a constrained optimization problem that finds the $L_p$-norm bounded noise $\bm\delta$ that minimizes the model's classification loss, while the outer minimization problem finds the parameters $\theta$ that also minimize the model's classification loss. 

\section{Experiment Results}

\begin{table}
\centering
\caption{The test accuracies (\%) of models trained on the clean training sets ($\mathcal{D}^{c}$) and their unlearnable ones ($\mathcal{D}^{u}$) for classification tasks.}
\label{table:1}
\resizebox{1.0\linewidth}{!}{
\begin{tabular}{ccccccc} 
\toprule
\multirow{2}{*}{Data} & \multirow{2}{*}{Clean} & AdvT                           & \multicolumn{2}{c}{EM}                                    & \multicolumn{2}{c}{Synthetic}                                     \\ 
\cmidrule(l){3-3}\cmidrule(l){4-5}\cmidrule(l){6-7}
                      &                        & $16/255$                       & $8/255$                 & $16/255$                        & $8/255$                        & $16/255$                         \\ 
\midrule
PathMNIST             & 87.8                   & \textbf{8.4($\downarrow$79.4)} & 13.5 ($\downarrow$74.3) & 19.6($\downarrow$68.2)          & 12.2($\downarrow$75.6)         & 16.9($\downarrow$70.9)           \\
DermaMNIST            & 72.0                   & 38.7($\downarrow$33.3)         & 17.1($\downarrow$54.9)  & \textbf{2.19($\downarrow$69.8)} & 11.5($\downarrow$60.5)         & 33.3($\downarrow$38.7)           \\
OctMNIST              & 69.6                   & 21.4($\downarrow$48.2)         & 25.0($\downarrow$44.6)  & \textbf{21.0($\downarrow$48.5)} & 22.7($\downarrow$46.9)         & 25.0($\downarrow$44.6)           \\
RetinaMNIST           & 52.0                   & 39.5($\downarrow$12.5)         & 13.0($\downarrow$39)    & 44.3($\downarrow$7.7)           & \textbf{8.3($\downarrow$43.7)} & 15.8($\downarrow$36.2)           \\
BreastMNIST           & 84.6                   & 44.9($\downarrow$39.7)         & 46.2($\downarrow$38.4)  & 73.1($\downarrow$11.5)          & 50.0($\downarrow$34.6)         & \textbf{37.2($\downarrow$47.4)}  \\
BloodMNIST            & 80.7                   & 19.6($\downarrow$61.1)         & 30.5($\downarrow$50.1)  & \textbf{17.4($\downarrow$63.3)} & 30.5($\downarrow$50.2)         & 27.8($\downarrow$52.9)           \\
TissueMNIST           & 54.4                   & 19.7($\downarrow$34.7)         & 17.4($\downarrow$37)    & \textbf{4.5($\downarrow$49.9)}  & 7.3($\downarrow$47.1)          & 7.1($\downarrow$47.3)            \\
OrganaMNIST           & 90.0                   & 81.0($\downarrow$9)            & 78.1($\downarrow$11.9)  & 67.3($\downarrow$22.7)          & 86.1($\downarrow$3.9)          & \textbf{59.4($\downarrow$30.6)}  \\
OrgancMNIST           & 90.7                   & 70.3($\downarrow$20.4)         & 72.2($\downarrow$18.5)  & 28.7($\downarrow$62)            & 74.8($\downarrow$15.9)         & \textbf{43.5($\downarrow$47.2)}  \\
OrgansMNIST           & 72.1                   & 51.7($\downarrow$20.4)         & 50.1($\downarrow$22)    & 27.3($\downarrow$44.8)          & 53.0($\downarrow$19.1)         & \textbf{22.8($\downarrow$49.3)}  \\
ChestMNIST            & 94.8                   & -                              & 78.7($\downarrow$16.1)  & \textbf{71.0($\downarrow$23.8)} & -                              & -                                \\
CheXpert              & 82.0                   & -                              & 76.7($\downarrow$5.3)   & \textbf{70.5($\downarrow$11.5)} & -                              & -                                \\
\bottomrule
\end{tabular}
}
\end{table}

We empirically verify the effectiveness of our method on multiple segmentation and classification datasets. On the classification tasks, as shown in Table \ref{table:1}, most of the test accuracy by the model trained on the protected dataset dropped rapidly and got worse when the protection perturbation radius increased. It is worth pointing out that the EM noise performs better when facing adversarial training on the DermaMNIST dataset. Surprisingly, the protected clean test accuracy is close to the result on the clean dataset for CheXpert and ChestMnist, indicating our protection does not work well for these tasks. 
For the segmentation task, Kvasir-SEG, as can be seen from Table \ref{table:2}, the IoU by the model trained on the protected dataset drops significantly to 0.00\%, which is perfect protection. 
Moreover, our method is effective under Adv. Training.

\begin{figure}
    \centering
    \includegraphics[width=0.8\linewidth]{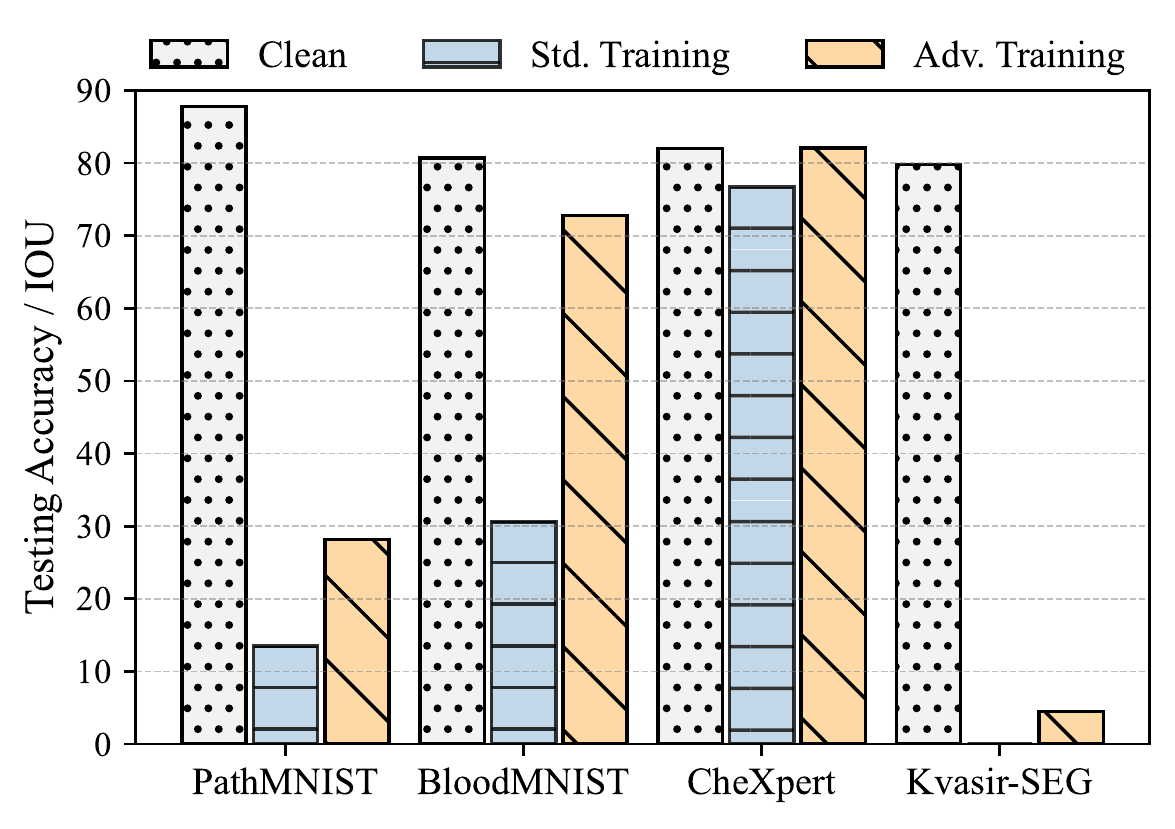}
    \caption{The testing accuracy / IOU of the model trained on data perturbed by EM noise under Std. training and Adv. training. 
    }
    \label{fig:fig2}
\end{figure}
\begin{table}
\caption{The IOU (\%) of models trained on the clean training sets ($\mathcal{D}^{c}$) and their unlearnable ones ($\mathcal{D}^{u}$) protected by EM Noise for segmentation tasks. $\epsilon_u$ and $\epsilon_a$ are the radius of noise perturbation and adversarial training.}
		 \label{table:2}	
   
\centering
\begin{tabular}{cccccc} 
\toprule
\multirow{2}{*}{Dataset} & \multirow{2}{*}{Clean} & \multicolumn{2}{c}{$\epsilon_u=8/255$} & \multicolumn{2}{c}{$16/255$}  \\ 
\cmidrule(lr){3-4}\cmidrule(lr){5-6}
                         &                        & $\epsilon_a=0$ & $4/255 $                & $0$ & $8/255$                   \\ 
\midrule
Kvasir-SEG              &     79.8    
              &            0.0 &    4.5               & 0.0                      & 1.2  \\
\bottomrule
\end{tabular}
\end{table}

\section{Discussion and Conclusion}
In this paper, we have explored the possibility of protecting biomedical data from unauthorized training using invisible noise. The result shows that all three types of noise work well, and the EM noise performs better in terms of availability. Furthermore, EM noise performs extremely well on the segmentation dataset. However, impaired effectiveness derived from adversarial training remains unsettled. An important future direction is to design more efficient and robust perturbation.
% The conclusion goes here.



\bibliographystyle{IEEEtranS}
\bibliography{ref}
%\begin{thebibliography}{99}
%\end{thebibliography}

\end{document}